\documentclass[a4paper]{cas-dc}

\usepackage[numbers]{natbib}

\usepackage{caption}
\usepackage{amsmath}
\usepackage{adjustbox}
\usepackage{multirow}
\usepackage{amssymb}
\usepackage{cleveref}

\usepackage[normalem]{ulem}
\usepackage{xcolor}

\def\tsc#1{\csdef{#1}{\textsc{\lowercase{#1}}\xspace}}
\tsc{WGM}
\tsc{QE}

\begin{document}
\let\WriteBookmarks\relax
\def\floatpagepagefraction{1}
\def\textpagefraction{.001}
\let\printorcid\relax 

\shorttitle{}    

\shortauthors{Y. Si et al.}

\title[mode = title]{SCSA: Exploring the Synergistic Effects Between Spatial and Channel Attention}  
\tnotemark[1]
\tnotetext[1]{This work was supported by the National Natural Science Foundation of China(62376252); Key Project of Zhejiang Provincial Natural Science Foundation(LZ22F030003).}

\author[1]{Yunzhong Si}
\ead{siyunzhong@zjnu.edu.cn} 

\author[1, 2]{Huiying Xu}
\ead{xhy@zjnu.edu.cn} 
\cormark[1]

\author[1, 2, 3]{Xinzhong Zhu}
\ead{zxz@zjnu.edu.cn}

\author[1]{Wenhao Zhang}
\ead{zwh2012918201@zjnu.edu.cn}

\author[1]{Yao Dong}
\ead{dongyao@zjnu.edu.cn}

\author[1]{Yuxing Chen}
\ead{cyx2001@zjnu.edu.cn}

\author[3]{Hongbo Li}
\ead{jason.li@geekplus.com}

\cortext[1]{Corresponding author. } 
\address[1]{College of Computer Science and Technology, Zhejiang Normal University, Jinhua, 321004, China}
\address[2]{Research Institute of Hangzhou Artificial Intelligence, Zhejiang Normal University, Hangzhou, 311231, China}
\address[3]{Beijing Geekplus Technology Co., Ltd, Beijing, 100101, China}

\begin{abstract}
Channel and spatial attentions have respectively brought significant improvements in extracting feature dependencies and spatial structure relations for various downstream vision tasks. The combined use of both channel and spatial attentions is widely considered beneficial for further performance improvement; however, the synergistic effects between channel and spatial attentions, especially in terms of spatial guidance and mitigating semantic disparities, have not yet been thoroughly studied. This motivates us to propose a novel Spatial and Channel Synergistic Attention module (SCSA), entailing our investigation toward the synergistic relationship between spatial and channel attentions at multiple semantic levels. Our SCSA consists of two parts: the Shareable Multi-Semantic Spatial Attention (SMSA) and the Progressive Channel-wise Self-Attention (PCSA). SMSA integrates multi-semantic information and utilizes a progressive compression strategy to inject discriminative spatial priors into PCSA's channel self-attention, effectively guiding channel recalibration. Additionally, the robust feature interactions based on the Channel-wise single-head self-attention mechanism in PCSA further mitigate the disparities in multi-semantic information among different sub-features within SMSA. We conduct extensive experiments on seven benchmark datasets, including classification on ImageNet-1K, object detection on MSCOCO, segmentation on ADE20K, and four other complex scene detection datasets. Our results demonstrate that our proposed SCSA not only surpasses the current plug-and-play state-of-the-art attention but also exhibits enhanced generalization capabilities across various task scenarios. The code and models are available at: \href{https://github.com/HZAI-ZJNU/SCSA}{https://github.com/HZAI-ZJNU/SCSA }.
\end{abstract}



\maketitle

\section{Introduction}
\label{sec:intro}
\noindent
Attention mechanisms, by enhancing representations of interest, facilitate the learning of more discriminative features and are widely used in redistributing channel relationships and spatial dependencies. Existing plug-and-play attention methods can be primarily categorized into three types: channel attention \cite{SENet, SKNet, GCT, ECANet, SCNet, FcaNet}, spatial attention \cite{GENet, NLNN, SASA, ViT}, and hybrid channel-spatial attention \cite{CBAM, BAM, TripleAttention, DualAttention, CA, SA, EMA, ELA}. Their focuses differ: channel attention focus on enhancing the extraction of key object features by adaptively weighting the importance of different channels, while spatial attention is tailored to augment critical spatial information. Spatial information represents semantic feature objects at the pixel level. Local spatial information captures low-semantic objects, such as fine details and textures, while global spatial information perceives high-semantic objects, such as overall shape. 

\begin{figure}[t]
    \centering
    \includegraphics[width=\linewidth]{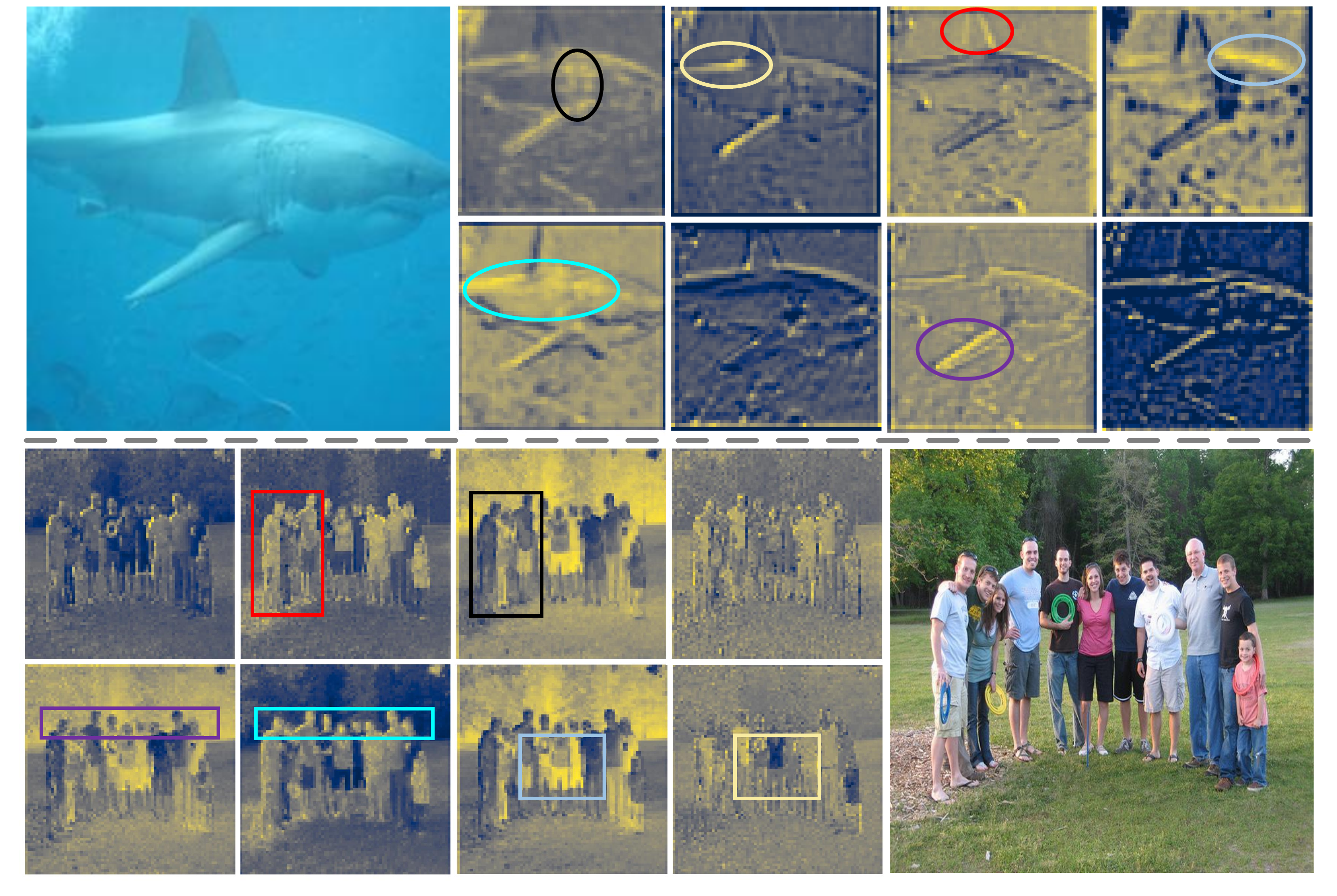}
    \caption{Visualization of several feature maps. Different box or circle colors highlight inherent spatial semantic disparities across specific parts in various feature maps.}
    \label{fig:featuremap}
\end{figure}

In models based on convolutional neural network (CNN) architectures, deep convolutional operators are commonly employed for feature extraction. The gradients generated by these operators can flow and propagate across different feature channels, facilitating the update of convolutional weights and effectively representing image features. However, numerous studies have shown that relying solely on the convolutional branch for feature extraction can lead to distorted gradient flows \cite{YOLOv9}, resulting in the loss of critical information or redundancy of similar features \cite{FasterNet, GhostNet, ShuffleNet, ShuffleNetV2}. To address this, several methods based on excitation and suppression mechanisms \cite{SENet} have been proposed, which focus feature learning on the most critical features for different tasks, thereby enhancing the model's representational capacity. In particular, CBAM \cite{CBAM} aggregates global spatial and channel information separately by chaining channel and spatial attention, but compressing all channel information leads to sharing across all spatial structures. This weakens the adaptability of spatial context to different feature maps. To overcome this, CPCA \cite{CPCA} introduces a channel-priority attention mechanism and depth-wise stripe convolutions, independently extracting spatial structures of each feature, significantly improving medical image segmentation. Furthermore, the EMA \cite{EMA} module, based on grouped attention and cross-spatial multi-scale interactions, effectively integrates spatial information of both long and short-range dependencies but overlooks inter-group feature interactions.

Although these hybrid attention mechanisms enhance representation learning, they overlook the inherent multi-semantic information across spatial and channel dimensions, as well as the interaction and disparity mitigation of multi-semantic features, which are crucial for fine-grained tasks such as detection and segmentation, thereby limiting the plug-and-play capability of these methods. As shown in \cref{fig:featuremap}, we analyze several feature maps of the image and observe that distinct spatial regions exhibit inherent semantic disparities and similarities, arising from the feature selectivity of different channels. 

Based on this insight, this raises the question of whether it is possible to leverage the inherent spatial semantic disparities across different feature channels to guide the learning of important features? Furthermore, given the presence of semantic disparities, how can we mitigate these multi-semantic differences and promote better fusion of multi-semantic information? 

Differing from the aforementioned methods, we explore solutions to the above issues from following three aspects: dimension decoupling, lightweight multi-semantic guidance, and semantic disparities mitigation, and propose a novel, plug-and-play Spatial and Channel Synergistic Attention (SCSA). Our SCSA is composed of a shareable Multi-Semantic Spatial Attention (SMSA) and a Progressive Channel-wise Self-Attention (PCSA) linked sequentially. Our study initially employs multi-scale, depth-shared 1D convolutions to extract spatial information at  various semantic levels from four independent sub-features. We utilize Group Normalization \cite{GN} across four sub-features to hasten model convergence while avoiding the introduction of batch noise and the interference of semantic information between different sub-features. Subsequently, we input the SMSA-modulated feature maps into PCSA, incorporating progressive compression and channel-specific single-head self-attention mechanisms. Our progressive compression strategy is designed to minimize computational complexity while preserving the spatial priors within SMSA, offering a practical trade-off. Moreover, our PCSA leverages an input-aware single-head self-attention mechanism to effectively explore channel similarities, thereby mitigating semantic disparities among different sub-features in SMSA and promoting information fusion. We conducted extensive experiments across four visual tasks and seven benchmark datasets, illustrating the effectiveness of the multi-semantic synergy applied in our SCSA. In summary, our contributions are as follows:

\begin{itemize}
    \item We identify two key limitations in existing plug-and-play attention mechanisms: 1) insufficient utilization of inherent multi-semantic spatial information to guide the extraction of key features along spatial and channel dimensions, and 2) inadequate handling of semantic disparities and interactions caused by multi-semantic information across feature maps.
    \item We propose the Spatial and Channel Synergistic Attention (SCSA), comprising the SMSA and PCSA modules. SMSA utilizes multi-scale depth-shared 1D convolutions to capture multi-semantic spatial information, enhancing both local and global feature representations. PCSA employs input-aware self-attention to refine channel features, effectively mitigating semantic disparities and ensuring robust feature integration across channels.
    \item Our proposed method outperforms other state-of-the-art plug-and-play attention mechanisms on multiple benchmarks, including ImageNet-1K for classification, MSCOCO for object detection, and ADE20K for segmentation, and demonstrates strong generalization capability across various complex scenarios, such as low-light and small-object benchmarks.
\end{itemize}

\begin{figure*}[t]
    \includegraphics[width=\linewidth]{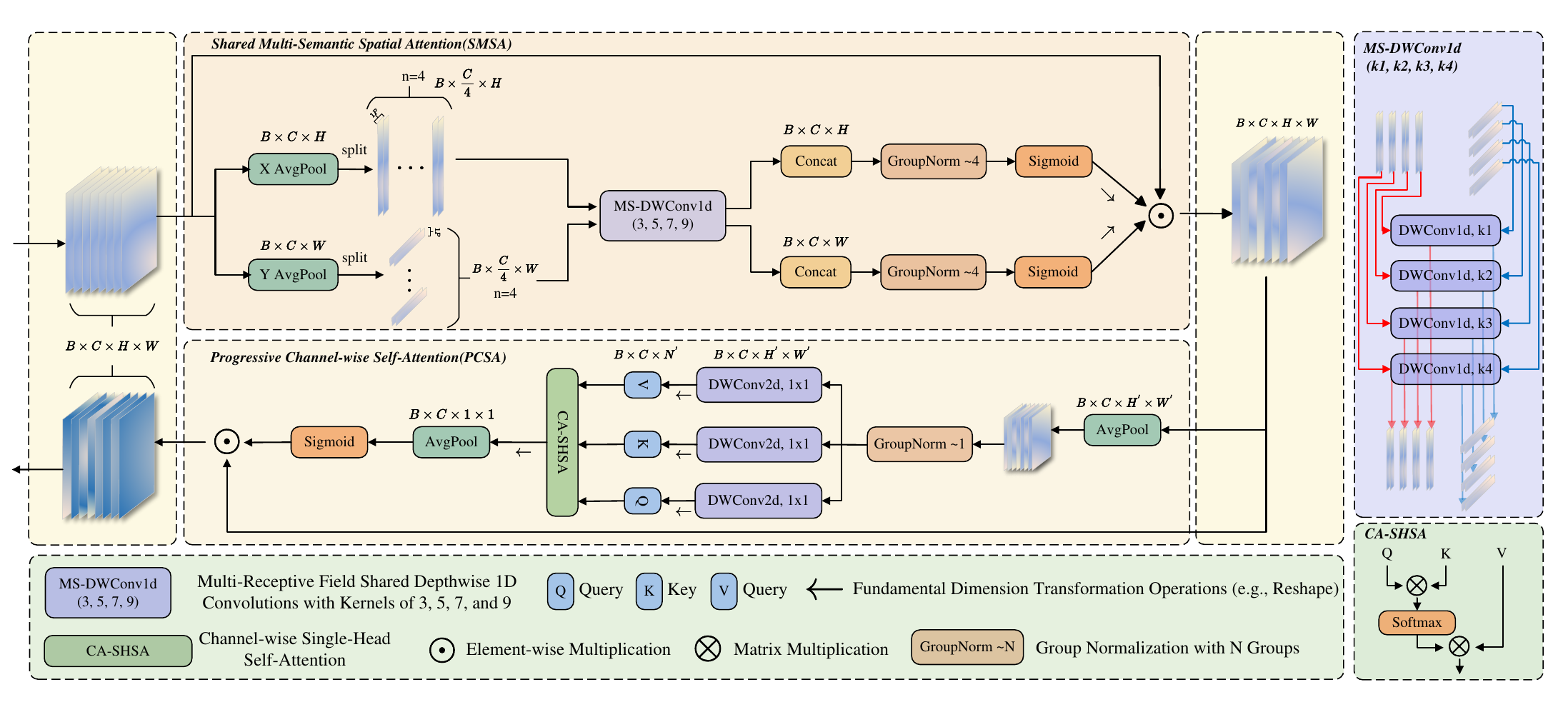}
    \caption{An illustration of our proposed SCSA, which uses multi-semantic spatial information to guide the learning of channel-wise self-attention. $B$ denotes the batch size, $C$ signifies the number of channels, and $H$ and $W$ correspond to the height and width of the feature maps, respectively. The variable $n$ represents the number of groups into which sub-features are divided, and $1P$ denotes a single pixel.}
    \label{fig:scsa}
\end{figure*}

\section{Related Work}
\label{sec:related}
\subsection{Multi-Semantic Spatial Information}
\noindent
Multi-semantic spatial structures incorporate rich category and contextual information. Effectively integrating global context and local spatial priors enables models to learn higher-quality representations from various perspectives. The InceptionNets \cite{InceptionV1, InceptionV2, InceptionV3, InceptionV4} pioneered a multi-branch approach, employing parallel vanilla convolutions of different sizes to capture varying receptive fields, significantly enhancing feature extraction capabilities. SKNet \cite{SKNet} incorporates multi-scale convolutions into channel attention, using the squeeze-and-excitation mechanism proposed by SENet \cite{SENet} to integrate spatial priors with varying receptive fields. Benefiting from the global contextual modeling ability, ViT \cite{ViT} employs MHSA to capture correlations at different spatial positions within distinct semantic sub-features, complemented by position embedding to compensate for spatial priors, achieving remarkable success in various downstream tasks. Currently, many studies develop efficient models \cite{ShuffleNet, ShuffleNetV2, FasterNet, RepViT} based on multi-semantic ideas, reducing parameters and computation for enhanced inference efficiency.  Mamba \cite{Mamba} introduces a selectable state space model using scanning mechanisms and GPU parallelism to model global contextual dependencies with linear time complexity. Additionally, VMamba \cite{VMamba} proposes a cross-scanning module that extends 1D sequence scanning to 2D image scanning, effectively capturing multi-semantic global context information from four directions.

\subsection{Attention Decomposition}
\noindent
Integrating attention mechanisms into mainstream backbones or feature fusion networks enhances fine-grained feature understanding and improves feature representation accuracy. However, it inevitably leads to increased memory usage and computational time. CA \cite{CA} and ELA \cite{ELA} perform unidirectional spatial compression along the height (H) and width (W) dimensions separately, preserving spatial structures in one direction while aggregating global spatial information in another, mitigating information loss from global compression. SA \cite{SA} and EMA \cite{EMA} reshape features into sub-features, reducing attention computation and parameters. However, the reshape operations they used in the high-dimensional B (batch size) and C (number of channels) constrained by GPU bandwidth can lead to expensive data transfers, significantly impacting inference speeds. CPCA \cite{CPCA} uses stripe convolutions in independent channels to reduce parameters in large-kernel convolutions. Recent studies also apply dimension decomposition in MHSA, with RMT \cite{RMT} applying MHSA separately across H and W dimensions to minimize computational costs.

In this study, we build upon the concept of attention decomposition and propose a lightweight guidance module that integrates multi-semantic spatial information. Additionally, we design a multi-semantic discrepancy mitigation module based on a progressive channel-wise single-head self-attention mechanism, aiming to explore a more optimized synergistic relationship between the spatial and channel dimensions.

\section{Method}
\label{sec:method}
\noindent
In this section, we begin by discussing the SMSA module, which explores the benefits of lightweight multi-semantic information guidance. Next, we introduce the PCSA module, which utilizes a progressive compression strategy and channel-wise self-attention to mitigate semantic disparities. The synergistic effects of multi-semantic guidance and semantic disparities mitigation motivate us to propose SCSA module. The overall architecture is shown in \cref{fig:scsa}.

\subsection{Shared Multi-Semantic Spatial Attention}
\label{sec:smsa}

\subsubsection{Spatial and Channel Decomposition. }
Decomposition techniques in neural network architectures substantially reduce the parameter count and computational overhead. Inspired by the structure of 1D sequences in Transformer \cite{Transformer}, in our work, we decompose the given input ${X \in \mathbb{R}^{B \times C \times H \times W}}$ along the height and width dimensions. We apply global average pooling to each dimension, thereby creating two unidirectional 1D sequence structures: $X_{H} \in \mathbb{R}^{B \times C \times W}$ and $X_{W} \in \mathbb{R}^{B \times C \times H}$. To learn varying spatial distributions and contextual relationships, we partition the feature set into $K$ identically sized, independent sub-features, $X_{H}^{i}$ and $X_{W}^{i}$, with each sub-feature having a channel count of $\frac{C}{K}$. In this paper, we set the default value $K=4$. The process of decomposing into sub-features is presented as follows:
{
\small
\begin{gather}
    X_{H}^{i} = X_{H}[:, (i-1) \times \frac{C}{K}: i \times \frac{C}{K}, :] \\ 
    X_{W}^{i} = X_{W}[:, (i-1) \times \frac{C}{K}: i \times \frac{C}{K}, :]
\end{gather}
}

\noindent
$X^{i}$ represents the $i$-th sub-feature, where $i \in [1, K]$. Each sub-feature is independent, facilitating efficient extraction of multi-semantic spatial information.

\subsubsection{Lightweight Convolution Strategies Across Disjoint Sub-features. }
After partitioning the feature maps into exclusive sub-features, we aim to efficiently capture distinct semantic spatial structures within each sub-feature. Inspired by extensive research on reducing feature redundancy \cite{ShuffleNetV2, GhostNet, FasterNet}, which reveal that such redundancy is likely due to intense interactions among features, we also observe varied spatial structures among features, as illustrated in \cref{fig:featuremap}. Based on these insights and aiming to enrich semantic information, enhance semantic coherence, and minimize semantic gaps, we apply depth-wise 1D convolutions with kernel sizes of 3, 5, 7, and 9 to four sub-features. Furthermore, to address the limited receptive field caused by decomposing features into H and W dimensions and applying 1D convolutions separately, we use lightweight shared convolutions for alignment, implicitly modeling the dependency between the two dimensions by learning consistent features across both. The ablation details regarding them are provided in \cref{tab:ablation}. The implementation process for extracting multi-semantic spatial information is defined as follows:
{
\small
\begin{gather}
    \tilde{X_{H}^{i}} = DWConv1d_{k_{i}}^{\frac{C}{K} \rightarrow \frac{C}{K} } (X_{H}^{i})\\
    \tilde{X_{W}^{i}} = DWConv1d_{k_{i}}^{\frac{C}{K} \rightarrow \frac{C}{K} } (X_{W}^{i})
\end{gather}
}

\noindent
$\tilde{X^{i}}$ represents the spatial structural information of the $i$-th sub-feature obtained after lightweight convolutional operations. $k_{i}$ denotes the convolution kernel applied to the $i$-th sub-feature.

After decomposing the independent sub-features and capturing the spatial information of different semantics, we need to construct the spatial attention map. Specifically, we concat distinct semantic sub-features and use Group Normalization (GN) \cite{GN} with $K$ groups for normalization. We opt for GN over the common Batch Normalization (BN) \cite{BN} because our study finds that GN is superior in distinguishing semantic differences among sub-features. GN allows for the independent normalization of each sub-feature without introducing batch statistical noise, effectively mitigating semantic interference between sub-features and preventing attention dilution. This design is validated by ablation studies shown in \cref{tab:ablation}. Finally, spatial attention is generated using a simple Sigmoid activation function, which activates and suppresses specific spatial regions. The computation of output features is as follows:
{
\small
\begin{gather}
    Attn_{H} = \sigma(GN_{H}^{K}(Concat(\tilde{X_{H}^{1}}, \tilde{X_{H}^{2}}, ..., \tilde{X_{H}^{K}}))) \\
    Attn_{W} = \sigma(GN_{W}^{K}(Concat(\tilde{X_{W}^{1}}, \tilde{X_{W}^{2}}, ..., \tilde{X_{W}^{K}}))) \\
    SMSA(X) = X_{s} = Attn_{H} \times Attn_{W} \times X
\end{gather}
}

\noindent
$\sigma(\cdot)$ denotes the Sigmoid normalization, while $GN_{H}^{K}(\cdot)$ and $GN_{W}^{K}(\cdot)$ represent GN with K groups along the H and W dimensions, respectively.

\subsection{Progressive Channel-wise Self-Attention}
\label{sec:pcsa}
\noindent
A prevalent approach to compute channel attention is through convolutional operations that explore dependencies among channels \cite{SENet, ECANet}. The use of convolution to model the similarities between features is somewhat non-intuitive and makes it difficult to effectively measure the similarity across different channels. Inspired by the significant advantages of the ViT \cite{ViT} in utilizing MHSA for modeling similarities among different tokens in spatial, we propose combining the Single-Head Self-Attention (SHSA) with modulated spatial priors from SMSA to compute inter-channel similarities. Moreover, to preserve and utilize the multi-semantic spatial information extracted by SMSA, and to reduce the computational cost of SHSA, we employ a progressive compression method based on average pooling, which serves as the \textbf{guidance} in our synergistic effects. Compared with modeling channel dependencies using common convolutional operations, PCSA exhibits stronger input perception capabilities and effectively leverages the spatial priors provided by SMSA to deepen learning. The implementation details of our PCSA are as follows:
{
\small
\begin{gather}
    X_{p} = Pool_{(7, 7)}^{(H, W) \rightarrow (H', W')}(X_{s}) \\
    F_{proj} = DWConv1d_{(1,1)}^{C \rightarrow C} \\
    Q = F_{proj}^{Q}(X_p), K = F_{proj}^{K}(X_p), V = F_{proj}^{V}(X_p) \\ 
    X_{attn} = Attn(Q,K,V) = Softmax(\frac{QK^{T}}{\sqrt{C}})V \\
    PCSA(X_{s}) = X_{c} = X_{s} \times \sigma(Pool_{(H', W')}^{(H', W') \rightarrow (1, 1)}(X_{attn}))
\end{gather}
}

\noindent
$Pool_{(k, k)}^{(H, W) \rightarrow (H', W')}(\cdot)$ denotes a pooling operation with a kernel size of $k \times k$ that rescales the resolution from $(H, W)$ to $(H', W')$. $F_{proj}(\cdot)$ represents the linear projection that generates the query, key, and value.

It's important to note that, unlike the MHSA in the ViTs where $Q, K, V \in \mathbb{R}^{B \times N \times C}$ with $N=HW$, in our PCSA's CA-SHSA, self-attention is computed along the channel dimension, with $Q, K, V \in \mathbb{R}^{B \times C \times N}$. Additionally, to fully interact with the different sub-features decomposed in SMSA, We select to implement a simpler single-head self-attention mechanism instead of the combination of multi-head self-attention with channel shuffling \cite{ShuffleNet}.

\subsection{Synergistic Effects}
\label{sec:scsa}
\noindent
The synergistic spatial and channel attention mechanisms aim to complement each other. In our work, we propose a novel concept of guiding channel attention learning through spatial attention. Drawing inspiration from the connection between CBAM \cite{CBAM} and CPCA\cite{CPCA}, we employ a similar serial structure to integrate our SMSA and PSCA modules, forming the Spatial and Channel Synergistic Attention (SCSA). The difference is that the spatial attention SMSA is applied first, followed by the channel attention PSCA. The former extracts multi-semantic spatial information from each feature, providing precise spatial priors for the latter; the latter refines the semantic understanding of the local sub-feature 
$X^i$ by leveraging the overall feature map $X$, thereby mitigating the semantic disparities caused by the multi-scale convolutions in the former. Additionally, unlike previous approaches \cite{SENet, CBAM, BAM, CA}, we do not employ channel compression, effectively preventing the loss of crucial features. Ultimately, our constructed SCSA is as follows:
{
\small
\begin{gather}
    SCSA(X) = PCSA(SMSA(X))
\end{gather}
}

\begin{table}[t]
    \centering
    \setlength{\tabcolsep}{1mm}
    \small
    \begin{tabular}{l|c|cc}
    \toprule
    \multirow{2}{*}{Ablations} & \multirow{2}{*}{Variants} & Throughput & Top-1\\
    &  & (imgs/s) & (\%) \\
    
    \midrule
     Baseline & \textbf{SCSA-50} & 2019 & \textbf{77.49} \\
    \cmidrule{1-4}
    
    \multirow{2}{*}{Macro Design} & w/o SMSA & 2217 & 77.21 \\
    &  w/o PCSA & 2155 & 77.44 \\
    \cmidrule{1-4}

    \multirow{2}{*}{Ordering} & PCSA Prior & 2005 & 77.20 \\
    & GN Prior & 2010 & 77.47 \\
    \cmidrule{1-4}
    
    \multirow{6}{*}{Micro Design} 
    & w/o Normalization in SMSA & 2010 & 77.15 \\
    & GN$\to$BN  & 1999 & 77.19 \\
    & GN$\to$LN  & 2005 & 77.20 \\
    & w/o PC & 1982 & 77.31 \\
    & w/ Multi-head + Shuffle & 2082 & 77.35 \\
    & Shared $\to$ Unshared & 1981 & 77.32 \\
    & Scaled: $ \sqrt{C} \to \sqrt{H * W}$ & 2001 & 77.34 \\
    \cmidrule{1-4}

    \multirow{3}{*}{Branch} & G1(3)& 2085 & 77.24 \\
    & G1(7) & 2063 & 77.17\\
    & G2(3,7) & 2040 & 77.32 \\

    \bottomrule

    \end{tabular}
     \caption{Ablation studies on the design strategy of SCSA, conducted at a 224x224 resolution, using the ImageNet-1K validation set. The abbreviation "PC" denotes progressive compression. $Gi(K_{1}, K_{2}, \ldots,  K_{i})$ denotes splitting $X$ into $i$ sub-features and applying a 1D convolution of size $K_{i}$ to each $i$-th sub-feature.}
    \label{tab:ablation}
\end{table}

\begin{figure}[t]
    \includegraphics[width=\linewidth]{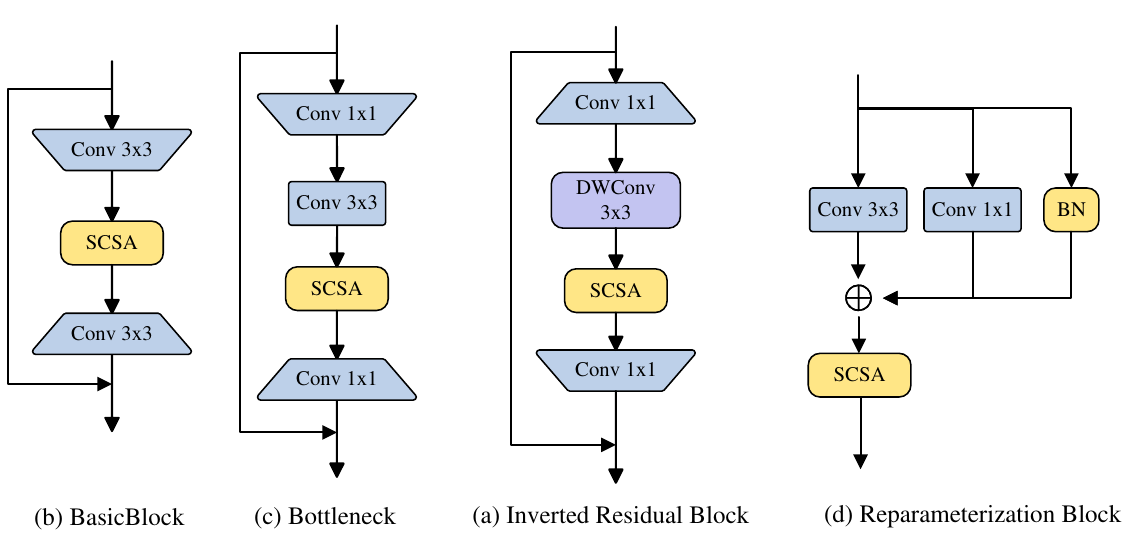}
    \caption{Main Module Structures with SCSA}
    \label{fig:block}
\end{figure}

\subsection{Integration of Attention Mechanisms}
\noindent
In our work, we integrate the proposed SCSA into different backbone networks to validate its effectiveness in enhancing feature extraction capabilities. As shown in \cref{fig:block}, the SCSA is integrated into four mainstream blocks: (a) and (b) represent blocks based on the ResNet \cite{ResNet} and its variant series \cite{ResNeXt}; (c) represents the inverted residual structure based on the MobileNet series \cite{MobileNetV2, MobileNetV3, MobileViT}; (d) represents the block structure of RepVGG \cite{RepVGG}, a representative of the reparameterization approach.

\begin{table*}[t]
    \centering
    \small
    \begin{tabular}{l|c|c|cc|c|cc}
    \toprule
    Backbones & Type & Methods & Params(M) & FLOPs(G) & Throughput(imgs/s) & Top-1(\%) & Top-5(\%) \\
    
    \midrule
    \multirow{11}{*}{ResNet-50} & -- & ResNet & 25.56 & 4.02 & 2433 & 76.39 & 93.09 \\
    \cmidrule(l{0pt}r{0pt}){2-8}
    & \multirow{3}{*}{Channel} & ECANet & 25.56 & 4.11 & 2109 & 77.05 & 93.43 \\
    & & SENet & 28.07 & 4.11 & 2077 & 77.23 & 93.56 \\
    & & FcaNet & 28.07 & 4.11 & 1905 & 77.29 & 93.64\\
    \cmidrule(l{0pt}r{0pt}){2-8}
    & \multirow{6}{*}{Hybrid} & CPCA & 27.40 & 4.87 & 1379 & 75.80 & 92.57\\
    & & CBAM & 28.07 & 4.12 & 1687 & 77.12 & 93.50 \\
    & & SANet & 25.56 & 4.12 & 1493 &77.12 & 93.64 \\
    & & ELA & 25.59 & 4.11 & 2233 & 77.25 & 93.52 \\
    & & CA & 25.69 & 4.11 & 2244 &77.37 & 93.52 \\
    & & EMA  & 25.57 & 4.18 & 1861 & 77.43 & \textbf{93.79} \\
    & & \textbf{SCSA(Ours)} & 25.62 & 4.12 & 2019 & \textbf{77.49} & 93.60 \\
    
    \midrule
    \multirow{7}{*}{ResNet-101} & -- & ResNet & 44.55 & 7.83 & 1588 & 77.76 & 93.81 \\
    \cmidrule(l{0pt}r{0pt}){2-8}
    & \multirow{3}{*}{Channel} & ECANet & 44.55 & 7.83 & 1408 &  78.32 & 93.99 \\
    & & SENet  & 49.30 & 7.84 & 1399 & 78.40 & 94.05 \\
    & & FcaNet & 49.29 & 7.84 & 1242 & 78.51 & 94.10\\
    \cmidrule(l{0pt}r{0pt}){2-8}
    & \multirow{3}{*}{Hybrid} & CBAM & 49.30 & 7.84 & 1118 & 78.09 & 94.07 \\
    & & CA & 44.80 & 7.84 & 1437 & 78.11 & 93.92 \\
    & & \textbf{SCSA(Ours)} & 44.68 & 7.85 & 1298 & \textbf{78.56} & \textbf{94.31}\\

    \midrule
    \multirow{5}{*}{MobileNetV2-1.0} & -- & MobileNetV2 & 3.51 & 0.31 & 6693 & 71.54 & 90.11 \\
    \cmidrule(l{0pt}r{0pt}){2-8}
    & \multirow{1}{*}{Channel} & ECANet & 3.51 & 0.31 & 5746 & 72.02 & 90.35  \\
    \cmidrule(l{0pt}r{0pt}){2-8}
    & \multirow{3}{*}{Hybrid}& CBAM & 4.07 & 0.32 & 4539 & 72.43 & 90.49 \\
    & & \textbf{SCSA(Ours)} & 3.63 & 0.34 & 2751 & \textbf{72.72} & \textbf{90.81}\\
    
    \midrule

    \multirow{5}{*}{RepVGG-A0} & -- & RepVGG & 9.11 & 1.52 & 6685 & 72.30 &  90.49 \\
    \cmidrule(l{0pt}r{0pt}){2-8}
    & \multirow{1}{*}{Channel} & ECANet & 9.11 & 1.52 & 5059 & 72.76 & 90.71   \\
    \cmidrule(l{0pt}r{0pt}){2-8}
    & \multirow{2}{*}{Hybrid} & CA & 9.35 & 1.52 & 3095 &  73.12 & 90.99  \\
    & & \textbf{SCSA(Ours)} & 9.18 & 1.53 & 2842 & \textbf{73.51} & \textbf{91.12} \\

    \midrule
        
    \multirow{2}{*}{Swin-T} & -- & Swin & 28.29 & 4.51 & 1523 & 80.83 & 95.49 \\
    & Hybrid & \textbf{SCSA(Ours)} & 28.36 & 4.52 & 1315 & \textbf{81.53} & \textbf{95.84}\\

    \bottomrule
    \end{tabular}
     \caption{Comparison of our proposed SCSA with other state-of-the-art attention mechanisms across multiple benchmark models at a 224x224 resolution on the ImageNet-1K validation set \cite{ImageNet1k}.}
    \label{tab:classify}
\end{table*}

\section{Experiments}
\label{sec:experiments}
\subsection{Experiments Settings}

\begin{table*}[t]
    \centering
    \small
    \begin{tabular}{l|c|cc|cccccc}
    \toprule
    Detectors & Methods & Params(M) & FLOPs(G) & AP(\%) & $AP_{50}$(\%) & $AP_{75}$(\%) & $AP_{S}$(\%) & $AP_{M}$(\%) & $AP_{L}$(\%) \\
    \midrule
    \multirow{13}{*}{Faster R-CNN} & ResNet-50 & 41.75 & 187.20 & 37.6 & 58.7 & 40.9 & 21.5 & 41.2 & 48.1\\
    & + FCA & 44.27 & 187.31 & 38.4 & 59.8 & 41.5 & 22.8 & 42.4 & 48.9 \\
    & + ECA & 41.75 & 187.20 & 38.5 & 60.0 & 41.4 & 22.6 & 42.6 & 49.4\\
    & + SE & 44.27 & 187.20 & 38.7 & 60.2 & 41.6 & \textcolor{blue}{\uline{23.2}} & 42.4 & 49.3\\
    & + CA & 41.89 & 187.21 & \textcolor{blue}{\uline{39.0}} & \textcolor{blue}{\uline{60.6}} & \textcolor{blue}{\uline{42.3}} & \textcolor{blue}{\uline{23.2}} & \textcolor{blue}{\uline{42.8}} & \textcolor{blue}{\uline{49.5}}\\
    & \textbf{+ SCSA(Ours)} & 41.81 & 187.35 & \textbf{39.3} & \textbf{60.6} & \textbf{42.8} & \textbf{23.2} & \textbf{43.1} & \textbf{50.2}\\
    \cmidrule{2-10}
    & ResNet-101 & 60.75 & 255.43 & 40.2 & 61.3 & 43.8 & 23.9 & 44.2 & 51.8\\
    & + FCA & 65.49 & 255.60 & 40.6 & 62.2 & 44.1 & 23.8 & 44.9 & 52.5 \\
    & + SE & 65.49 & 255.44 & 40.8 & 62.2 & 44.4 & \textbf{24.9} & 44.7 & 53.0 \\
    & + ECA & 60.75 & 255.43 & 40.9 & \textcolor{blue}{\uline{62.4}} & 44.3 & 24.2 & \textcolor{blue}{\uline{45.0}} & 53.0 \\
    & + CA & 60.99 & 255.45 & \textcolor{blue}{\uline{41.1}} & 62.2 & \textcolor{blue}{\uline{44.8}} & 24.1 & \textcolor{blue}{\uline{45.0}} & \textcolor{blue}{\uline{53.5}}\\
    & \textbf{+ SCSA(Ours)} & 60.88 & 255.74 & \textbf{41.5} & \textbf{62.9} & \textbf{45.4} & \textcolor{blue}{\uline{24.6}} & \textbf{45.3} & \textbf{53.7} \\
    \midrule

    \multirow{12}{*}{Cascade R-CNN} & ResNet-50 & 69.40 & 214.84 & 40.3& 58.9& 43.8& 22.5& 43.8& 52.8\\
    & + FCA & 71.91 & 214.94 & 41.3 & 60.2 & 44.6 & 24.1 & 44.9 & 53.7 \\
    & + SE & 71.91 & 214.84 & 41.4 & 60.2 & 44.9 & 24.5 & 44.7 & 54.0\\
    & + CBAM & 71.91 & 214.87 & 41.4 & 60.2 & 45.0 & 24.5 & 44.6 & \textcolor{blue}{\uline{54.3}}\\
    & + ECA & 69.40 & 214.84 & \textcolor{blue}{\uline{41.7}} & \textcolor{blue}{\uline{60.7}} & \textcolor{blue}{\uline{45.2}} & \textbf{24.7} & \textcolor{blue}{\uline{45.4}} & \textcolor{blue}{\uline{54.3}}\\
    & \textbf{+ SCSA(Ours)} & 69.46 & 214.99 & \textbf{42.1}& \textbf{61.4} & \textbf{45.7} & \textcolor{blue}{\uline{24.6}} & \textbf{45.5} & \textbf{54.3} \\
    \cmidrule{2-10}
    & ResNet-101 & 88.39 & 283.07 & 42.6 & 61.1 & 46.6 & 24.9 & 46.7 & 55.7\\
    & + SE & 93.13 & 283.08 & 43.2 & 62.3 & 47.2 & 25.8 & 47.1 & 56.2\\
    & + FCA & 93.13 & 283.24 & 43.4 & 62.5 & 47.6 & 25.5 & 47.3 & 56.8 \\
    & + ECA & 88.39 & 283.07 & 43.7 & 62.7 & 47.5 & 25.5 & \textcolor{blue}{\uline{47.7}} & 56.8\\
    & + CA & 88.64 & 283.09 & \textcolor{blue}{\uline{43.8}} &\textcolor{blue}{\uline{62.8}} &  \textcolor{blue}{\uline{48.0}} & \textcolor{blue}{\uline{26.0}} & 47.6 & \textcolor{blue}{\uline{57.4}}\\
    & \textbf{+ SCSA(Ours)} & 88.52 & 283.38 & \textbf{44.2} & \textbf{63.1} & \textbf{48.2} & \textbf{26.0} & \textbf{48.2} & \textbf{57.5} \\
    \midrule

    \multirow{13}{*}{RetinaNet} & ResNet-50 & 37.97 & 214.68 & 36.5& 55.5& 39.1& 20.2 & 40.1 & 48.1 \\
    & + FCA & 40.48 & 214.78 & 37.3 & \textcolor{blue}{\uline{57.2}} & 39.3 & 21.6 & 40.9 & 49.0\\
    & + SE & 40.49 & 214.68 & 37.4 & 57.0 & 40.0 & 21.5 & \textcolor{blue}{\uline{41.3}} & 49.0\\
    & + ECA & 37.97 & 214.68 & 37.5 & \textcolor{blue}{\uline{57.2}} & 39.8 & 21.5 & 41.1 & \textcolor{blue}{\uline{49.5}}\\
    & + CBAM & 40.49 & 214.71 & \textcolor{blue}{\uline{37.6}} & 57.0 & \textcolor{blue}{\uline{40.2}} & \textcolor{blue}{\uline{22.0}} & \textbf{41.6} & 48.7 \\
    & \textbf{+ SCSA(Ours)} & 38.03 & 214.83 & \textbf{37.9}& \textbf{57.6} & \textbf{40.2} & \textbf{22.5} & \textcolor{blue}{\uline{41.3}} & \textbf{49.7} \\
    \cmidrule{2-10}
    & ResNet-101 & 56.96 & 282.91 & 39.3 & 58.7 & 41.9 & 22.8 & 43.5 & 51.8\\
    & + SE & 61.71 & 282.92 & 39.8 & 59.9 & 42.2 & 22.9 & 43.8 & 52.1\\
    & + FCA & 61.70 & 283.08 & 39.9 & 60.0 & 42.4 & 22.9 & \textbf{44.6} & 52.4\\
    & + CA & 57.21 & 282.93 & 40.2 & 60.0 & \textcolor{blue}{\uline{43.0}} & 23.2 & 44.3 & \textcolor{blue}{\uline{52.8}}\\
    & + ECA & 56.96 & 282.91 & \textcolor{blue}{\uline{40.3}} & \textcolor{blue}{\uline{60.4}} & 42.9 & \textcolor{blue}{\uline{23.4}} & 44.2 & 52.7\\
    & \textbf{+ SCSA(Ours)} & 57.09 & 283.22 & \textbf{40.5} & \textbf{60.8} & \textbf{43.6} & \textbf{23.7} & \textcolor{blue}{\uline{44.3}} & \textbf{53.1}\\

    \bottomrule
    \end{tabular}
       \caption{Comparison of the performance of different attention mechanisms for object detection on the MSCOCO validation set \cite{MSCOCO}, utilizing models such as Faster R-CNN \cite{Faster-RCNN}, Cascade R-CNN \cite{Cascade-RCNN}, and RetinaNet \cite{RetinaNet}. All models were fine-tuned using the "$1\times$" schedule with optimal results in bold and suboptimal results underlined in blue.}
    \label{tab:det}
\end{table*}

\noindent
In this section, we first introduce the experimental details. Next, we conduct experiments on four visual tasks, comparing our proposed SCSA with other state-of-the-art attention mechanisms. Following this, in  \cref{subsec:ablation}, we perform a comprehensive ablation study on our meticulously designed SCSA from four different perspectives.

\subsubsection{Datasets. }
We validate the effectiveness of our method across four visual tasks. For the image classification, we select the widely used ImageNet-1K \cite{ImageNet1k} dataset. In the object detection, we employ several challenging detection datasets, including MSCOCO \cite{MSCOCO}, Pascal VOC \cite{VOC0712}, VisDrone \cite{VisDrone2019}, and ExDark \cite{ExDark}. For semantic segmentation and instance segmentation, we selected the widely used ADE20K \cite{ADE20k} and MSCOCO \cite{MSCOCO} benchmarks. 

We are keen to explore whether attention mechanisms can be more effectively applied to various complex scene tasks. While previous attention research \cite{SENet, ECANet, SA, EMA, CBAM, ELA} has shown good performance on widely used benchmarks (e.g., ImageNet-1K \cite{ImageNet1k}, MSCOCO \cite{MSCOCO}), the effectiveness in dense, low-light, and small-object scenes remains uncharted. Therefore, we conduct more experiments using representative benchmarks in \cref{tab:otherdet}: the small-object dataset VisDrone \cite{VisDrone2019}, low-light dataset ExDark \cite{ExDark}, infrared automotive dataset FLIR-ADAS v2 \cite{FLIR-ADAS-V2}, and general dataset Pascal VOC \cite{VOC0712}.

\subsubsection{Metrics. }
We use Top-1 and Top-5 metrics to measure image classification, Average Precision (AP) to evaluate object detection, and report Parameter Count (Params) and Floating Point Operations Per Second (FLOPs), and throughput to measure performance. For semantic segmentation, we employ the mean Intersection over Union (mIoU).

\subsubsection{Implementation Details. }
\label{subsec:imp-detail}

To evaluate our proposed SCSA on ImageNet-1K \cite{ImageNet1k}, we select four mainstream backbone networks based on CNN and Transformer architectures, including ResNet \cite{ResNet}, MobileNetV2 \cite{MobileNetV2}, RepVGG \cite{RepVGG} and Swin \cite{Swin}. Specifically, we follow the parameter configurations in the original papers \cite{ResNet, MobileNetV2, RepVGG, Swin}, except for the batch size and learning rate. Since all classification models are trained on a single NVIDIA RTX 4090 GPU, we adjust the batch size and learning rate according to the linear scaling rule \cite{goyal2017accurate, you2017scaling}. For ResNet \cite{ResNet}, RepVGG \cite{RepVGG}, and Swin \cite{Swin}, the batch size is uniformly set to 128, with the learning rates scaled down to 0.05, 0.05, and 0.000125, respectively. When training MobileNetV2 with our SCSA, we use the batch size and learning rate from ECANet \cite{ECANet}, set to 96 and 0.045, respectively. Notably, to enhance training efficiency, we employ Automatically Mixed Precision(AMP) training.

We evaluate our SCSA on MSCOCO \cite{MSCOCO} using Faster R-CNN \cite{Faster-RCNN}, Mask R-CNN \cite{Mask-RCNN}, Cascade R-CNN \cite{Cascade-RCNN}, and RetinaNet \cite{RetinaNet}. These detectors are implemented using the MMDetection \cite{MMDetection} toolboxes with default settings. All models are trained using an SGD optimizer with a momentum of 0.9 and a weight decay of 1e-4, with a batch size of 2 per GPU, over a total of 12 epochs. Faster R-CNN, Mask R-CNN and Cascade R-CNN started with a learning rate of 0.0025, while RetinaNet starts at 0.00125. The learning rates for all models are decreased by a factor of 10 at the 8th and 11th epochs. We fine-tuned the model on MSCOCO \cite{MSCOCO} for 12 epochs using a single NVIDIA H800 GPU and reported comparative results on validation set. Building on the above configurations, we further evaluate the proposed SCSA method’s detection performance and generalization capability on Pascal VOC \cite{VOC0712}, as well as in complex scenarios such as VisDrone \cite{VisDrone2019}, ExDark \cite{ExDark}, and FLIR-ADAS V2 \cite{FLIR-ADAS-V2}.

We further validate our method on ADE20K \cite{ADE20k} with the UperNet \cite{UperNet} for semantic segmentation. Following common practices \cite{DeepLab, InceptionNeXt}, we utilize the MMSegmentation \cite{MMSegmentation} toolboxes, set the batch size to 16, and conduct 80k training iterations. All models are trained using an SGD optimizer with an initial learning rate of 0.01, a momentum of 0.9, and a weight decay of 5e-4. We also conduct training and inference using a single NVIDIA H800 GPU.

All models are trained with the default random seed 0.

\subsection{Image Classification}
\label{subsec:img-class}
\noindent
We compare our SCSA against other state-of-the-art attention mechanisms, including SENet \cite{SENet}, CBAM \cite{CBAM}, ECANet \cite{ECANet}, FcaNet (FCA) \cite{FcaNet}, CA \cite{CA}, SANet \cite{SA}, EMA \cite{EMA}, CPCA \cite{CPCA} and ELA \cite{ELA}. As shown in \cref{tab:classify}, our SCSA achieved the highest Top-1 accuracy across networks of different scales, with negligible parameter count and computational complexity. Within hybrid architectures, our method's throughput based on ResNet is second only to CA and ELA, but it offer a better balance of accuracy, speed, and model complexity with a moderate model width. Integrating the SCSA method into the MobileNetV2 architecture significantly improves model accuracy. Although SCSA is more lightweight in terms of parameter count (3.63M vs. 4.07M, -0.44M), its multi-branch structure encounters a sharp increase in channel dimensions within the inverted residual blocks, which leads to a reduction in throughput. Notably, integrating the proposed SCSA method into advanced models like RepVGG \cite{RepVGG} and the spatial self-attention-based Swin \cite{Swin} still yields notable accuracy improvements of 1.21\% and 0.70\%, respectively, effectively demonstrating the adaptability of our attention mechanism across different model architectures.

\subsection{Object Detection}
\label{subsec:secobj-detect}

\subsubsection{Results on MSCOCO. }
We evaluate various attention mechanisms on MSCOCO to verify the effectiveness of our approach in dense detection scenario. We use ResNet-50 and ResNet-101 as the backbone and FPN \cite{FPN} as the feature fusion network. As shown in \cref{tab:det}, our method outperforms other state-of-the-art attention methods across various detectors, model sizes, and object scales. For Faster R-CNN \cite{Faster-RCNN}, our SCSA improves by 1.7\% and 1.3\% in terms of AP compared to the original ResNet-50 and ResNet-101, respectively. Compared to other plug-and-play attention modules, including CBAM \cite{CBAM}, FCA \cite{CBAM}, ECA \cite{ECANet}, and CA \cite{CA}, SCSA demonstrates superior performance, achieving gains of 0.4\% to 1.0\% on the Cascade R-CNN \cite{Cascade-RCNN} detector. Moreover, it consistently excels in detecting targets across a range of scales, confirming its robust adaptability to multi-scale features.

\begin{table}[t]
    \centering
    \small
    \setlength{\tabcolsep}{1mm}
    \begin{tabular}{l|c|ccc|ccc}
    \toprule
    \multirow{2}{*}{Datasets} & \multirow{2}{*}{Methods} & \multicolumn{3}{c}{ResNet-50} & \multicolumn{3}{c}{ResNet-101}\\
    \cmidrule{3-8}
    & & AP & $AP_{50}$ & $AP_{75}$ & AP & $AP_{50}$ & $AP_{75}$\\
    \midrule
    \multirow{6}{*}{Pascal VOC} & - & 50.7 & 81.9 & 55.7 & 54.3 & 83.8 & 61.0\\
    & +SE & 50.2 & 81.9 & 54.1 & 53.7 & 83.6 & 60.1\\
    & +ECA & 50.7 & 82.2 & 55.0 & 54.4 & \textcolor{blue}{\uline{84.4}} & 60.5\\
    & +FCA & 50.8 & 82.0 & 55.2 & 53.7 & 83.7 & 59.6\\
    & +CA & \textcolor{blue}{\uline{51.8}} & \textcolor{blue}{\uline{82.5}} & \textcolor{blue}{\uline{56.5}} & \textcolor{blue}{\uline{55.4}} & 84.2 & \textcolor{blue}{\uline{61.5}}\\
    & \textbf{+SCSA} & \textbf{53.0} & \textbf{83.0} & \textbf{58.0} & \textbf{55.5} & \textbf{84.6} & \textbf{61.8}\\
    \midrule
    \multirow{6}{*}{VisDrone2019} & - & 22.1 & 37.3 & 23.1 & 23.1 & 38.5 & 24.5\\
    & +SE & 21.6 & 36.7 & 22.4  & 22.1 & 37.6 & 23.1\\
    & +FCA & 21.9 & 37.1 & 22.7 & 22.4 & 38.0 & 22.8\\
    & +ECA & 21.9 & 37.3 & 22.7 & 22.6 & 38.3 & 22.9 \\
    & +CA & \textcolor{blue}{\uline{22.8}} & \textcolor{blue}{\uline{38.3}} & \textcolor{blue}{\uline{23.9}} & \textbf{23.5} & \textcolor{blue}{\uline{39.2}} & \textbf{24.4}\\
    & \textbf{+SCSA} & \textbf{22.9} & \textbf{38.7} & \textbf{24.0} & \textcolor{blue}{\uline{23.3}} & \textbf{39.2} & \textcolor{blue}{\uline{24.2}}\\
    \midrule
    \multirow{6}{*}{ExDark} & - & 39.2 & 71.4 & 38.6 & 42.4 & 74.9 & 43.4\\
    & +ECA & 37.9 & 70.7 & 37.2 & 42.4 & 75.1 & 42.8\\
    & +SE & 38.3 & 71.1  & 37.1 & 41.8 & 74.8 & 42.0\\
    & +FCA & 38.3 & 71.4 & 37.6 & 41.9 & 75.0 & 42.4\\
    & +CA & \textcolor{blue}{\uline{39.5}} & \textcolor{blue}{\uline{72.2}} & \textcolor{blue}{\uline{39.8}} & \textbf{43.2} & \textcolor{blue}{\uline{75.6}} & \textbf{45.4}\\
    & \textbf{+SCSA} & \textbf{40.2} & \textbf{73.2} & \textbf{40.0} & \textcolor{blue}{\uline{43.0}} & \textbf{75.6} & \textcolor{blue}{\uline{44.9}}\\
    \midrule
    \multirow{6}{*}{FLIR-ADAS v2} & - & \textcolor{blue}{\uline{24.7}} & 42.2 & 25.5 & \textbf{26.3} & \textbf{44.6}  & \textbf{28.0} \\
    & +CA & 24.2 & 42.2 & 25.0 & \textcolor{blue}{\uline{25.5}} & \textcolor{blue}{\uline{43.7}} & \textcolor{blue}{\uline{26.8}}\\
    & +FCA & 24.4 & 41.5 & 25.8 & 24.7 & 42.0 & 25.9\\
    & +SE & 24.5 & \textbf{42.5}  & 25.5  & 25.2 & 42.9 & 26.0 \\
    & +ECA & 24.6 & 41.9  & \textcolor{blue}{\uline{25.6}} & 25.3 & 42.8 & 25.9\\
    & \textbf{+SCSA} & \textbf{24.8} & \textcolor{blue}{\uline{42.3}} & \textbf{26.1} & 25.4 & 43.2 & 26.2 \\
    \bottomrule
    \end{tabular}
      \caption{Comparison of our SCSA, based on ResNet-50 and ResNet-101, with other attention mechanisms for object detection performance across four different datasets.}
    \label{tab:otherdet}
\end{table}

\subsubsection{Results on Infrared, Low-Light, and Small Target Detection. }
As shown in \cref{tab:otherdet}, it is gratifying to see that proposed
SCSA performs better across these benchmarks \cite{ExDark, VisDrone2019, VOC0712, FLIR-ADAS-V2} compared to other counterparts, further demonstrating the robustness of our strategy in maintaining channel dimensions and the synergistic concept of multi-semantic information. Notably, our results indicate that there are still some limitations in the application of attention mechanisms on long-tail datasets, such as FLIR-ADASv2 \cite{FLIR-ADAS-V2}, has led to minimal performance gains and even declines. This may be due to the attention mechanism’s squeeze-and-excitation strategy being illsuited for handling imbalanced distributed data, resulting in a focus on high-frequency categories while neglecting the learning of low-frequency ones.

\begin{table}[t]
    \centering
     \small
    \begin{tabular}{lccc}
    \toprule
    \multirow{2}{*}{Methods} & \multicolumn{3}{c}{UperNet} \\
    \cmidrule{2-4}
    & Params(M) & FLOPs(G) & mIoU(\%) \\
    \midrule

    ResNet-50 & 64.10 & 1895 & 40.20 \\
    + CBAM & 66.62 & 1895 & 39.62 \\
    + CPCA &  65.94 &  1927 & 39.68 \\
    + SE & 66.62 & 1895 & 39.94 \\
    + SA & 64.10 & 1895 & 40.01 \\
    + ECA & 64.10 & 1895 & 40.46 \\ 
    + FCA & 66.61 & 1895 & 41.09\\
    \textbf{+ SCSA(Ours)} & 64.16 &  1895 & \textbf{41.14} \\
    \midrule
    ResNet-101 & 83.09 & 2051 & 42.74 \\
    + CBAM & 87.84 & 2051& 41.65\\
    + ECA & 83.09 & 2051 & 42.63\\
    + SE & 87.84 & 2051 & 42.66\\
    + FCA & 87.83 & 2051 & 43.22 \\
     \textbf{+ SCSA(Ours)} & 83.22 & 2051 & \textbf{43.76} \\
    \bottomrule
    \end{tabular}
     \caption{Comparison of our method, based on the UperNet model, with other attention mechanisms for semantic segmentation performance on the ADE20K benchmark.}
    \label{tab:semantic}
\end{table}

\begin{table}[t]
    \centering
    \small
    \begin{tabular}{lcccccc}
    \toprule
    \multirow{2}{*}{Methods} & \multicolumn{6}{c}{Mask R-CNN} \\
    \cmidrule{2-7}
    & AP & $AP_{50}$ & $AP_{75}$& $AP_{S}$ & $AP_{M}$ & $AP_{L}$ \\
    \midrule
    ResNet50 & 34.8 & 55.9 & 36.9 & 16.4 & 37.4 & 50.2\\
    + CBAM & 35.4 & 56.9 & 37.6 & \textcolor{blue}{\uline{17.4}} & 38.3 & 50.6 \\
    + ECA & 35.5 & 57.6 & 37.6 & 16.6 & 38.4 & \textbf{52.0}\\
    + FCA & 35.5 & 57.2 & 37.6 & 17.1 & 38.6 & 51.3\\
    + SE & 35.7 & 57.3 & 38.1 & \textbf{17.7} & 38.6 & 50.9\\
    + SA & 35.7 & \textcolor{blue}{\uline{57.7}} & 38.0 & 17.2 & \textcolor{blue}{\uline{38.7}} & 51.5\\
    + CA & \textcolor{blue}{\uline{35.8}} & 57.5 & \textcolor{blue}{\uline{38.2}} & 16.9 & 38.5 & 51.7\\
    \textbf{+ SCSA} & \textbf{36.1} & \textbf{58.4} & \textbf{38.3} & 17.2 & \textbf{39.1} & \textcolor{blue}{\uline{51.9}}\\
    \bottomrule
    \end{tabular}
    \caption{Comparison of our method, based on the Mask R-CNN, with other attention mechanisms for instance segmentation performance on the MSCOCO validation set \cite{MSCOCO}.}
    \label{tab:instance}
\end{table}

\subsection{Segmentation}
\label{subsec:sem-seg}
\noindent
We also test its performance in semantic segmentation on ADE20K \cite{ADE20k} and instance segmentation on MSCOCO \cite{MSCOCO}. We conduct extensive comparative experiments based on the UperNet \cite{UperNet} network. As shown in  \cref{tab:semantic} and \cref{tab:instance}, our SCSA significantly outperforms other attention methods. Specifically, SCSA improves performance by 0.94\% and 1.02\% in terms of mIoU on ResNet-50 and ResNet-101, respectively, while other methods only achieve improvements of 0.1\% to 0.2\%, and some even fall below the baseline model. Meanwhile, SCSA achieves a 0.3\% to 0.7\% increase in terms of AP in instance segmentation tasks, surpassing other counterparts. These results demonstrate that our method, based on multi-semantic spatial information, performs well in pixel-level tasks.

\subsection{Ablation study}
\label{subsec:ablation}
\noindent
As shown in \cref{tab:ablation}, we apply SCSA to ResNet-50, constructing SCSA-50 as the baseline on ImageNet-1K \cite{ImageNet1k} for ablation studies across four aspects.

\subsubsection{Macro Design. } We validate the SMSA and PCSA modules separately, and both show significant improvements in accuracy compared to ResNet-50. With SMSA, guided by multi-semantic information, the Top-1 accuracy improved significantly by 1.05\%, while PCSA, which mitigates multi-semantic disparities and promotes channel interaction, increased the accuracy by 0.82\%. Without progressive compression in PCSA, accuracy drops by 0.18\%, and this is primarily because, after direct global spatial compression, PCSA cannot leverage the discriminative spatial priors provided by SMSA for its computations.

\subsubsection{Ordering. } 
Our study primarily aims to explore whether the inherent multi-semantic spatial information across the spatial and channel dimensions can effectively guide the learning of channel features. To further demonstrate the benefits of this "guidance," we swapped the order of PCSA and SMSA. Interestingly, the Top-1 accuracy dropped by 0.29\%, which further validates our previous hypothesis that spatial attention can guide channel feature learning, thereby confirming the effectiveness of guiding with multi-semantic information. 

Although normalization helps reduce the impact of data noise and accelerates model convergence \cite{WeightNorm}, the placement of normalization within SMSA may yield varying effects. Backbone networks based on MHSA \cite{Transformer, ViT} typically use Layer Normalization (LN) \cite{LN} before attention computation, whereas some plug-and-play attention modules either omit normalization layers \cite{ECANet, CPCA} or apply them beforehand \cite{SA, EMA}. To explore the necessity and optimal placement of normalization in SMSA, we conduct experiments by placing normalization before attention and by removing it. Results shown in \cref{tab:ablation} indicate that normalization is essential for the attention mechanism, though its specific position has a minor impact. Pre-normalization aids in handling variations among input features and improves training stability, but may cause loss of feature details, reducing attention sensitivity to fine-grained information. In contrast, applying normalization after attention calculation can mitigate noise but may also diminish the model's focus on important features. Ultimately, based on accuracy results, we opted to place normalization after attention calculation in SMSA.

\begin{figure}[t]
    \centering
    \includegraphics[width=\linewidth]{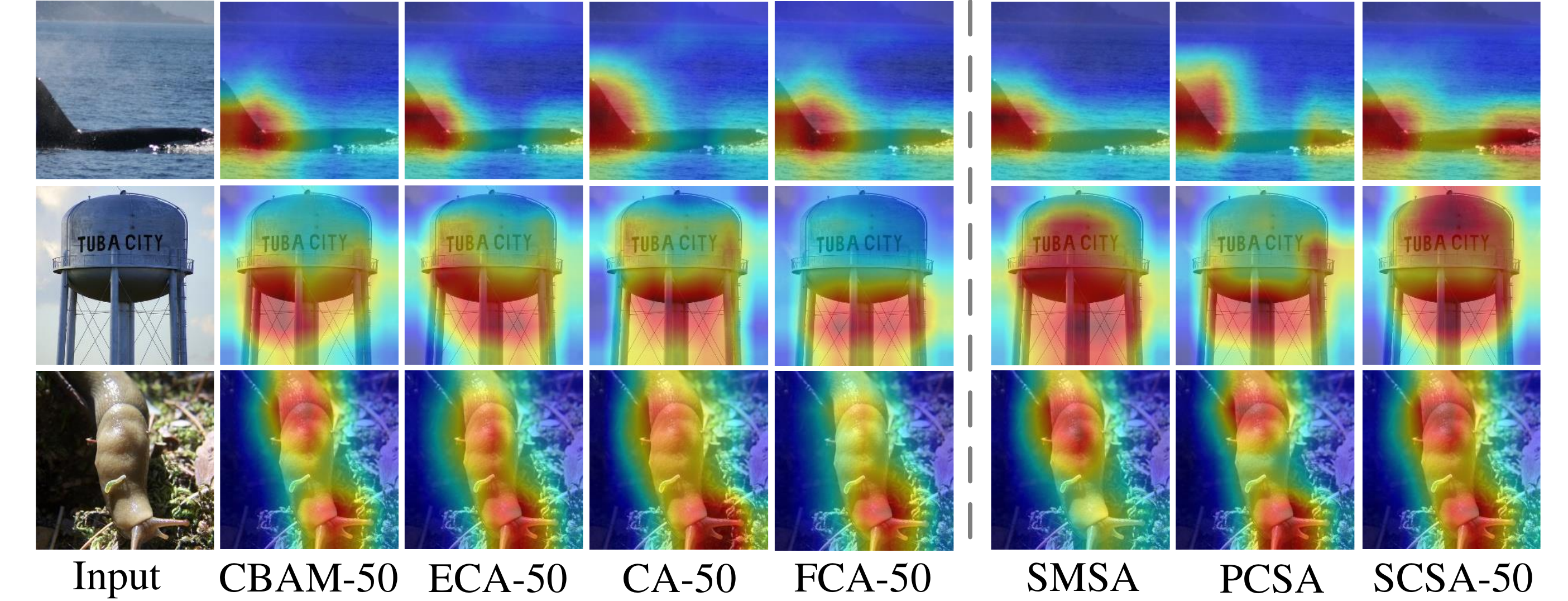}
    \caption{Comparative attention visualizations for 'layer 4.2' across multiple models, generated using samples randomly selected from different categories of the ImageNet-1K validation set, through Grad-CAM \cite{GradCAM}.}
    \label{fig:vis}
\end{figure}

\subsubsection{Micro Design. } Following the experimental analysis above, which confirms the importance of normalization layers in attention calculation, we consider whether group normalization is more suitable for extracting multi-semantic information from multiple sub-features in the proposed SCSA. To investigate this, we perform ablation studies comparing popular normalization methods in deep neural networks (DNNs), such as BN \cite{BN} and LN \cite{LN}. The results show that when GN is sequentially replaced with BN and LN, both accuracy and inference speed decrease, with Top-1 accuracy dropping to 77.19\% and 77.20\%, respectively. These declines are attributed to GN's superior ability to preserve the independence of semantic patterns among sub-features, thereby minimizing semantic interference. Conversely, BN's sensitivity to batch size can introduce statistical noise \cite{BNWeak} when processing multi-semantic information. LN, by normalizing along the channel dimension and capturing information across all features, may disrupt the distinct semantic patterns that SMSA’s multi-scale convolutions extract from individual sub-features. These ablation suggest that GN may be a more suitable choice in convolution layers that involve multiple semantics. Furthermore, the decline in accuracy and increase in parameters with unshared convolutions further validate the effectiveness of using shared convolutions to consistently learn and model features dependencies across the H and W dimensions.

Additionally, when replacing the single attention head in PCSA with multi-head and channel shuffle operation \cite{ShuffleNet}, performance decrease from 77.49\% to 77.35\%. This phenomenon is primarily attributed to the strong inter-channel interactions facilitated by the single head, which effectively alleviate semantic disparities produced in SMSA. To validate the shared convolutional learning on 1D sequences decomposed along the Height and Width dimensions, we compare it to non-shared convolutional learning. Results show a 0.17\% accuracy drop, reduced throughput, and increased parameters and FLOPs due to more convolutional operators. This confirms that shared learning across dimensions captures complementary features, enhancing model expressiveness.

\begin{figure*}[t]
    \includegraphics[width=\linewidth]{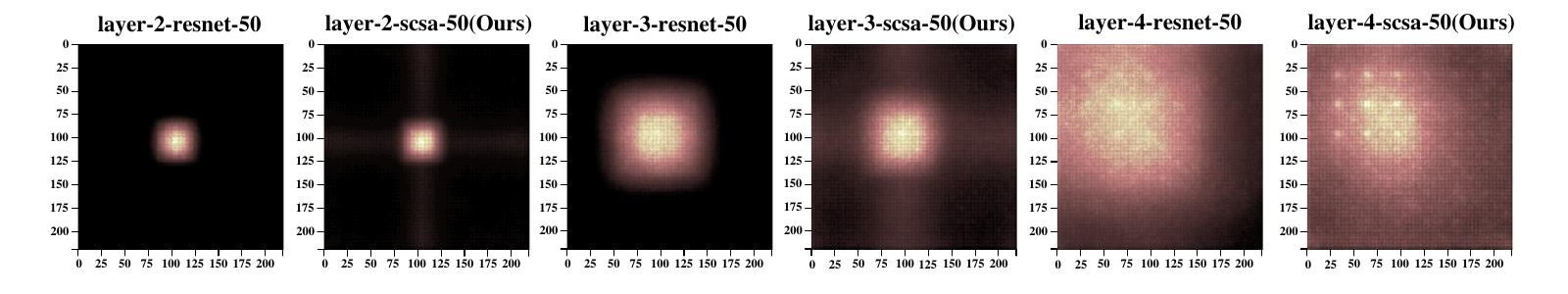}
    \caption{Comparison of effective receptive fields (ERFs). Our SCSA provides a larger effective receptive field compared to the baseline, and the effect becomes more pronounced as the layers deepen.}
    \label{fig:erf}
\end{figure*} 

\subsubsection{Branch. } 
The richness of semantic feature capture in SMSA is determined by the number of branches and the convolution kernel sizes used in each branch. Each branch is designed to learn a distinct sub-features. Reducing the number of branches weakens the module's ability to extract inherent multi-semantic features. To assess the impact of capturing different semantic features on model performance, we conducted experiments with varying branch numbers and convolution kernel sizes. As shown in the "Branch" section of \cref{tab:ablation}, the accuracy of the dual-branch structure surpasses that of the single-branch, while the four-branch structure further outperforms the dual-branch. This results supports the effectiveness of our multi-branch, multi-scale structure in capturing diverse semantic patterns across sub-features, thereby enhancing the model’s representational capacity. As the number of branches increases, the model’s memory access overhead also rises, resulting in a decrease in inference speed.

\section{Visualization and Analysis}
\label{sec:analysis}
\subsection{Visualization of Attention}
\noindent
We evaluate the effectiveness of our method in mitigating semantic disparities and enhancing consistency by ensuring appropriate attention to key regions. As shown in \cref{fig:vis}, compared to other state-of-the-art attention mechanisms, our SCSA distinctly focuses on multiple key regions, significantly reducing critical information loss while providing rich feature information. We also visualize the components of SCSA, including the SMSA and PCSA modules. In the absence of the PCSA module to address semantic disparities, the distribution of activation intensity remains inadequately balanced. Without the SMSA module for guiding multi-semantic spaces, the focus on important regions may be limited. 

\subsection{Visualization of Effective Receptive Field}
\noindent
As depicted in \cref{fig:erf}, leveraging the spatial structure of multi-semantic modeling, our SCSA  has achieved a broader perceptual area. A larger effective receptive field (ERF) is beneficial for the network to utilize rich contextual information for collective decision-making, which is one of the important factors for performance improvement. To verify that the performance of our method benefits from a larger ERF, we randomly sample 300 images of different categories from the ImageNet-1K validation set \cite{ImageNet1k}, measure the contribution of each pixel on the original image to the center point of the output feature maps of the third and fourth stages of the model, and quantify the range of the ERF with the gradient values weighted and normalized. The visualization results demonstrate that as the network layers deepen, the ERF of our SCSA becomes increasingly evident, confirming our hypothesis and the effectiveness of our method.

\subsection{Computational Complexity}
\noindent
Given an input ${X \in \mathbb{R}^{B \times C \times H \times W}}$, a pooling size of $P \times P$, and a depth-wise convolutional kernel size of $K \times K$, we sequentially consider the impact of dimension decoupling, depth-shared 1D convolutions, normalization, progressive compression, and channel-wise self-attention, which collectively constitute the SCSA module. For simplicity of observation, we ignore the coefficients. The computational complexities of SCSA are:
{
\small
\begin{align}
    \Omega(SCSA) &= \mathcal{O}(HC+WC) + \mathcal{O}(KHC + KWC) \notag\\
                 &\quad+ \mathcal{O}(HWC) + \mathcal{O}(P^{2}H^{'}W^{'}C + H^{'}W^{'}C) \notag\\
                 &\quad+ \mathcal{O}(H^{'}W^{'}C + H^{'}W^{'}C^{2})
\end{align}
}

\noindent
$H^{'}$ and $W^{'}$ denote the height and width, respectively, of the intermediate feature map produced by the progressive compression operation.

We observe that when the model width (i.e., the number of channels, $C$) is moderate, $\Omega(SCSA)$ scales linearly with the length of the input sequence. This indicates that our SCSA can perform inference with linear complexity when the model width is moderate.

\begin{figure*}[t]
    \centering
    \includegraphics[width=\linewidth]{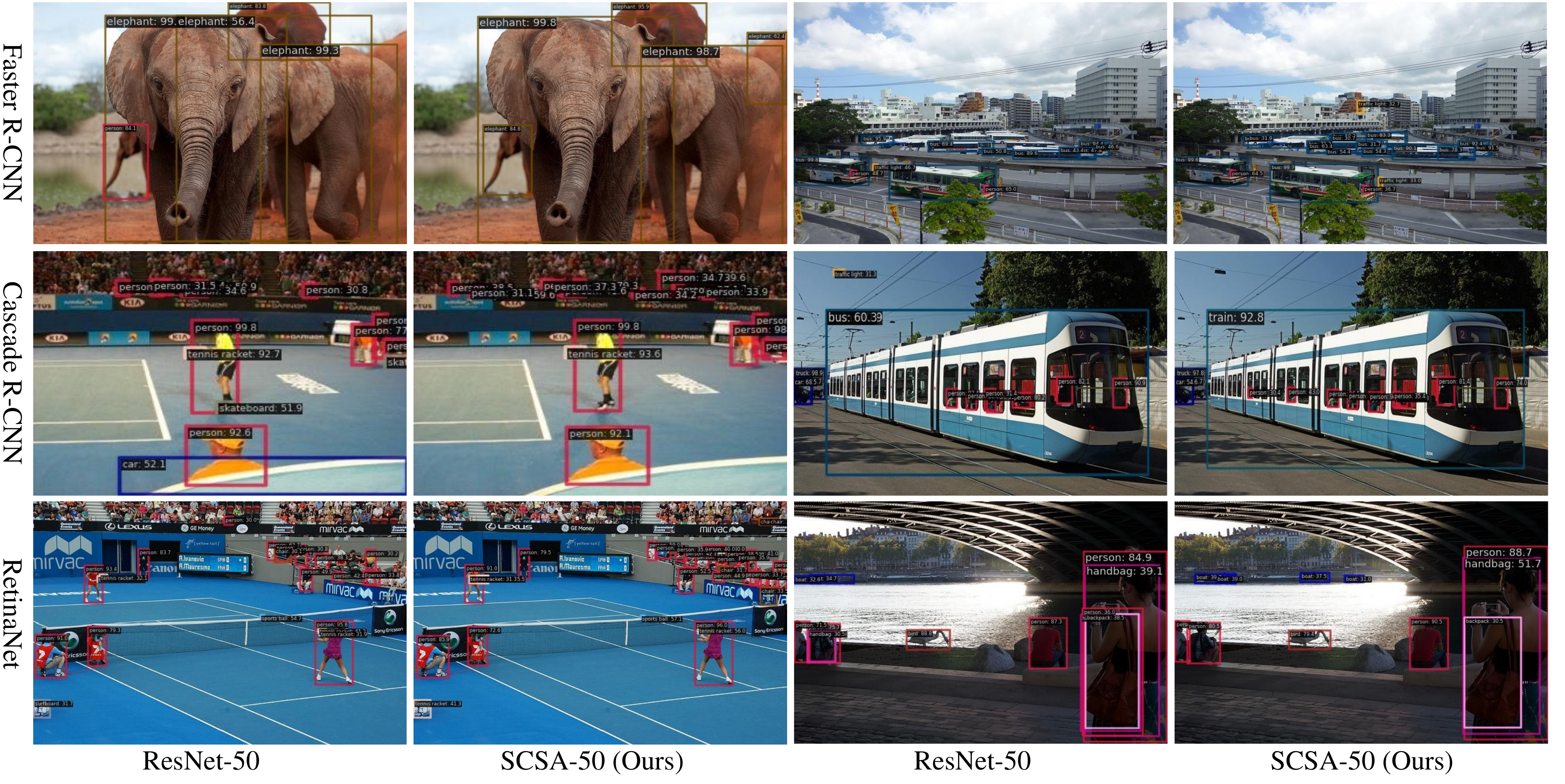}
    \caption{Detection results are visualized on Faster R-CNN \cite{Faster-RCNN}, Cascade R-CNN \cite{Cascade-RCNN}, and RetinaNet \cite{RetinaNet} by respectively selecting two random samples from MSCOCO validation set \cite{MSCOCO} and comparing our SCSA with the ResNet-50 \cite{ResNet} baseline to demonstrate the effectiveness of our method.}
    \label{fig:det-res}
\end{figure*}

\subsection{Inference Throughput Evaluation}
As demonstrated in \cref{tab:ablation} and \cref{tab:classify}, we evaluate the throughput of SCSA's individual components in ablation experiments and compare the throughput across various benchmark models using different attention mechanisms. We conduct our experiments using a GeForce RTX 4090 GPU at a resolution of 224x224, with a batch size of 32 to simulate real-world applications and maximize GPU utilization. To minimize variability, we repeat 100 times for each attention mechanism and report the average inference time. Specifically, As illustrated in \cref{tab:classify}, although SCSA is slightly slower than pure channel attention, it outperforms most hybrid attention mechanisms, including CBAM, SANet, EMA, and CPCA, and achieves the highest accuracy.

\subsection{Qualitative Results of Object Detection}
\noindent
 As shown in \cref{fig:det-res}, our method demonstrates superior performance in challenging scenarios, including obstruction, dense environments, clusters of small objects, and low-light conditions.

\begin{figure}[t]
   \centering
    \includegraphics[width=\linewidth]{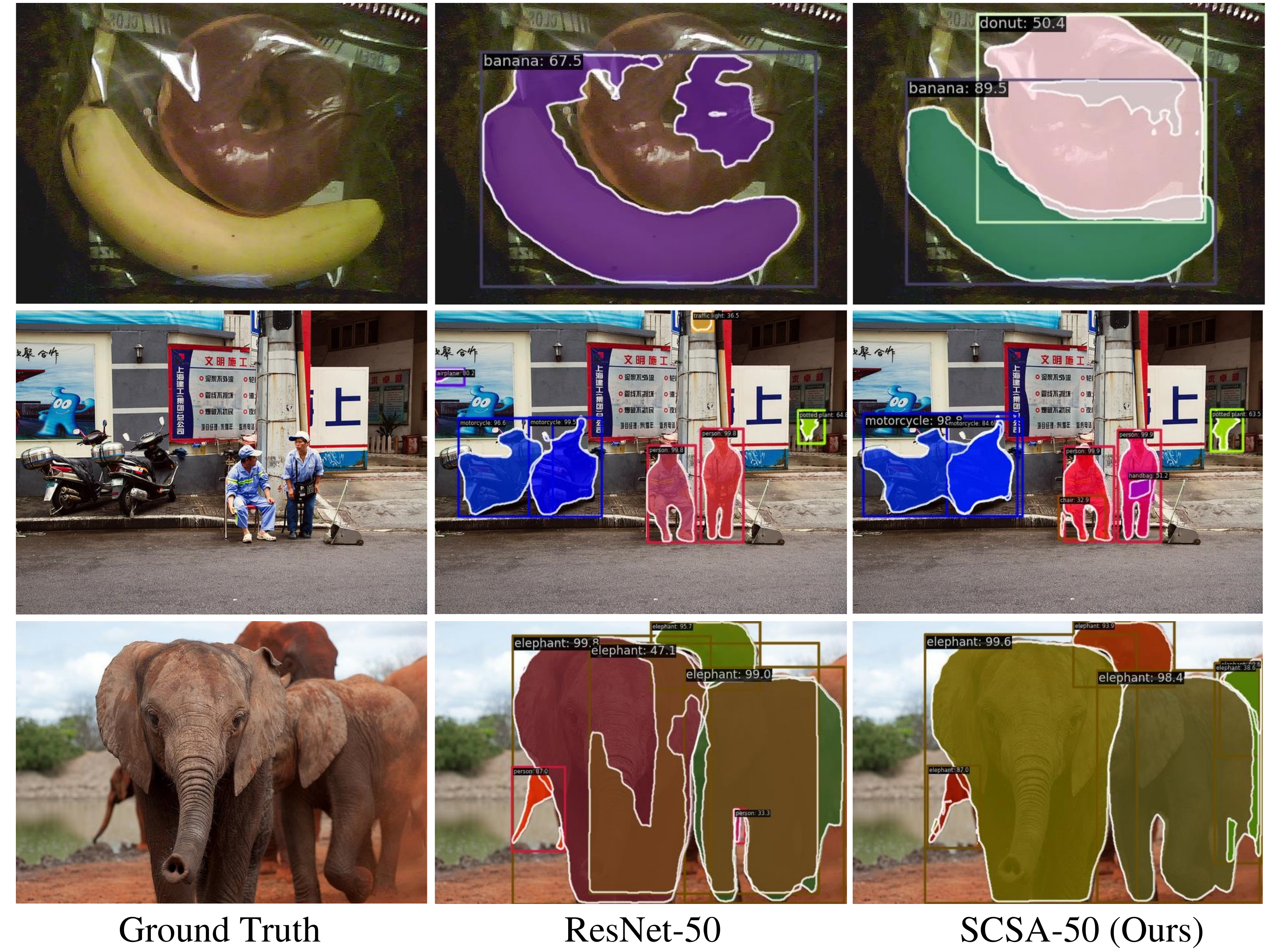}
    \caption{Visualization of instance segmentation results using the Mask R-CNN \cite{Mask-RCNN}. Each instance depicted in a distinct color.}
    \label{fig:instance-seg-res}
\end{figure}

\subsection{Qualitative Results of Instance Segmentation}
\noindent
 As shown in \cref{fig:instance-seg-res}, our method segments obscured and overlapping objects more comprehensively and accurately, achieving higher confidence scores. These results underscore the benefits of our method in leveraging multi-semantic information to better perceive the contextual space of relevant objects.

\begin{figure}[t]
   \centering
    \includegraphics[width=\linewidth]{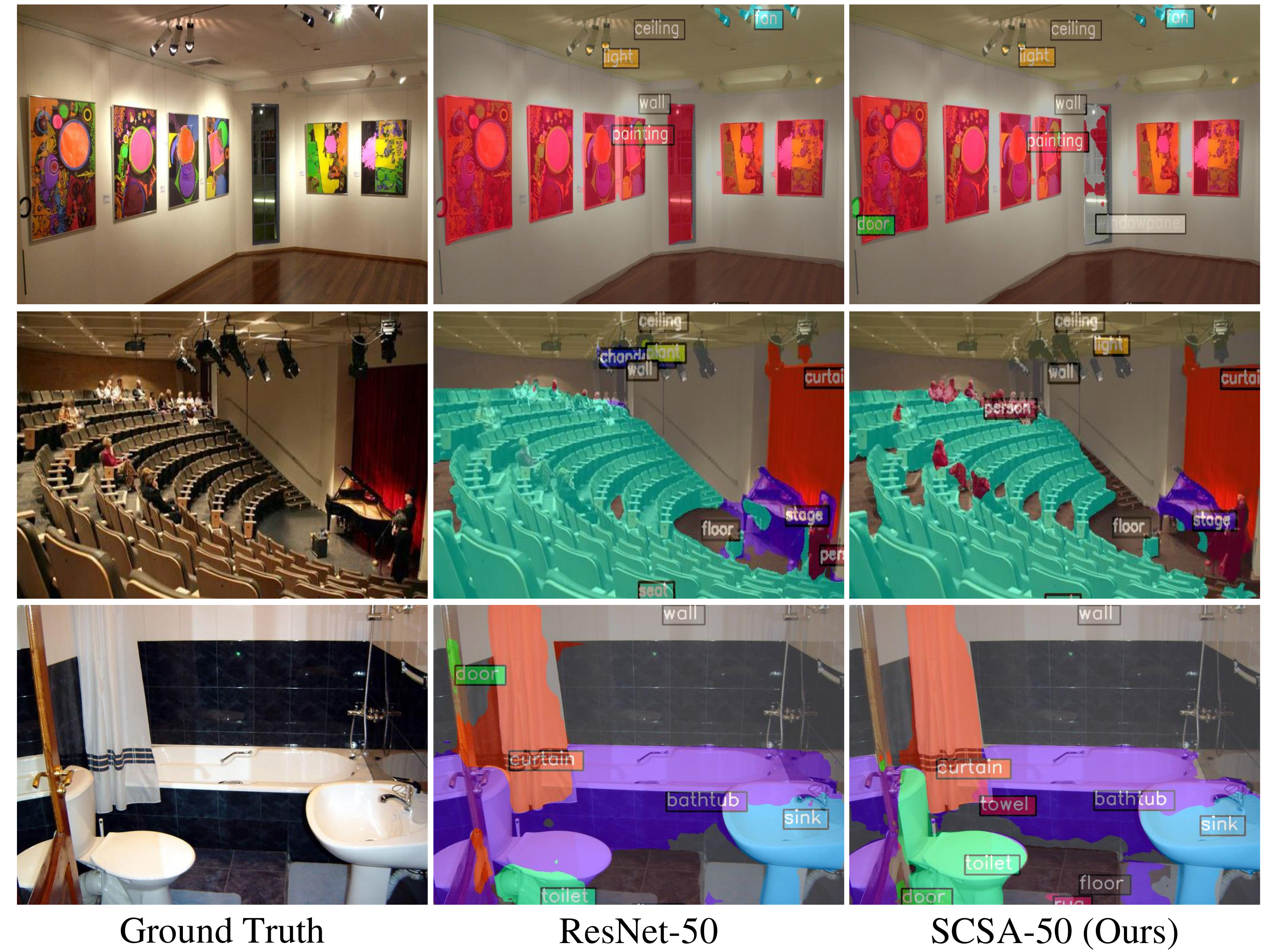}
    \caption{Visualization of semantic segmentation results using the UperNet \cite{UperNet} on ADE20K \cite{ADE20k}.}
    \label{fig:semantic-seg-res}
\end{figure}

\subsection{Qualitative Results of Semantic Segmentation}
\noindent
 It can be observed from \cref{fig:semantic-seg-res} that our method significantly improves the segmentation of objects that overlap and are semantically adjacent, effectively distinguishing between scenarios such as spectators seated on chairs and toilets near bathtubs.

\section{Limitations}\label{sec:limitations}
We demonstrated that our SCSA, a plug-and-play synergistic attention method, excels in image classification, object detection, and instance and semantic segmentation. Although we are committed to exploring the synergistic effects across various dimensions and have empirically validated the effectiveness of leveraging multi-semantic spatial information to guide channel recalibration and enhance feature interactions for mitigating semantic disparities, inference latency remains a significant challenge in real-world deployment. Our approach achieves an optimal balance of model parameters, accuracy, and inference speed at an appropriate model width. However, at larger widths, the primary bottleneck in inference speed is the use of depth-wise convolutions and branching within the construction of a mutli-semantic spatial structure, which have low FLOPS, frequently access memory, and exhibit low computational density \cite{FasterNet, MobileOne, EfficientMod}. We believe that the positioning and quantity of these plug-and-play attention modules should be optimized based on specific tasks and scenarios to ensure peak performance. In the future, we will investigate more lightweight and faster plug-and-play attention mechanisms, exploring the synergistic relationships across different dimensions.

\section{Conclusion}
\label{sec:conclusion}
\noindent
In this study, we analyze the limitations of most plug-and-play attention methods in leveraging the inherent multi-semantic information of features across spatial and channel dimensions, as well as the challenges posed by semantic disparities. To address these issues, we propose a novel plug-and-play Spatial and Channel Synergistic Attention (SCSA) mechanism, which incorporates dimension decoupling, lightweight multi-semantic guidance, and semantic disparity mitigation. SCSA leverages multi-semantic spatial attention to guide the learning of diverse channel features, followed by single-head self-attention in the channel dimension to alleviate semantic disparities and promote semantic interaction. Extensive experiments demonstrate that SCSA consistently outperforms state-of-the-art attention mechanisms on widely used benchmarks, showing enhanced performance and robust generalization capabilities. We hope our work encourages further exploration of synergistic properties across multiple dimensions in various domains.











\bibliographystyle{cas-model2-names}

\bibliography{cas-refs}

\end{document}